\documentclass[lettersize,journal]{IEEEtran}
\usepackage{amsmath,amsfonts}
\usepackage{algorithmic}
\usepackage{algorithm}
\usepackage{array}
\usepackage[caption=false,font=normalsize,labelfont=sf,textfont=sf]{subfig}
\usepackage{textcomp}
\usepackage{stfloats}
\usepackage{url}
\usepackage{verbatim}
\usepackage{graphicx}
\usepackage{multirow}
\usepackage{booktabs}
\usepackage[normalem]{ulem}
\usepackage{cite}
\begin{document}

\title{Synthetic Data Augmentation for Enhanced Chicken Carcass Instance Segmentation }

\author{Yihong Feng, Chaitanya Pallerla, Xiaomin Lin, Pouya Sohrabipour Sr, Philip Crandall, Wan Shou, Yu She, Dongyi Wang 
\thanks{Yihong Feng, Pouya Sohrabipour Sr, Dongyi Wang are with the Department of Biological and Agricultural Engineering, University of Arkansas, Fayetteville, AR 72701 USA (e-mail: yihongf@uark.edu, pouyas@uark.edu, dongyiw@uark.edu). 

Chaitanya Pallerla, Philip Crandall are with the Department of Food Science, University of Arkansas, Fayetteville, AR 72701 USA (e-mail: pallerla@uark.edu, crandal@uark.edu). 

Xiaomin Lin is with Department of Electrical Engineering, University of South Florida, Tampa, FL 33620 USA (e-mail: xlin2@usf.edu). 

Wan Shou is with Department of Mechanical Engineering, University of Arkansas, Fayetteville, AR 72701 USA (e-mail: wshou@uark.edu).

Yu She is with Department of Industrial Engineering, Purdue University, West Lafayette, IN 47907, USA (e-mail: yushe@purdue.edu).
This paper was produced by the IEEE Publication Technology Group. They are in Piscataway, NJ.}
}

\markboth{Submitted for reviewing}%
{Shell \MakeLowercase{\textit{et al.}}: A Sample Article Using IEEEtran.cls for IEEE Journals}


\maketitle

\begin{abstract}
The poultry industry has been driven primarily by broiler chicken production and has grown into the world’s largest animal protein sector. Automated detection of chicken carcasses on processing lines is vital for quality control, food safety, and operational efficiency in slaughterhouses and poultry processing plants. However, developing robust deep learning models for tasks like instance segmentation in these fast-paced industrial environments is often hampered by the need for laborious acquisition and annotation of large-scale real-world image datasets. 

To this end, we present the first pipeline generating photo-realistic, automatically labeled synthetic images of chicken carcasses under varying poses. We also introduce a new benchmark dataset containing 300 annotated real-world images, curated specifically for poultry segmentation research. Using these datasets, this study investigates the efficacy of synthetic data  and automatic data annotation to enhance the instance segmentation of chicken carcasses, particularly when real annotated data from the processing line is scarce. A small real dataset (60 images of chicken carcasses) with varying proportions of synthetic images were evaluated in prominent instance segmentation models: YOLOv11-seg, Mask R-CNN (with R50 and R101 backbones), and Mask2Former. Results show that synthetic data significantly boosts segmentation performance for chicken carcasses across all models. YOLOv11-seg consistently achieved the highest accuracy. Notably, models with greater capacity (R101) and transformer-based architectures (Mask2Former) derived greater benefits, particularly with larger volumes of synthetic data. Furthermore, model-specific optimal ratios of synthetic-to-real data were observed. This research underscores the value of synthetic data augmentation as a viable and effective strategy to mitigate data scarcity, reduce manual annotation efforts, and advance the development of robust AI-driven automated detection systems for chicken carcasses in the poultry processing industry.
\end{abstract}


\section{Introduction}
\IEEEPARstart{O}{ver} the past two decades, driven by increased demand in established and emerging markets, poultry has become the most widely consumed animal protein worldwide\cite{USDAERSPoultry}. Moreover, the poultry industry is projected to remain the world’s largest meat exporting sector over the next decade\cite{USDAERSPoultry}. From 2001 to 2021, the rapid expansion of poultry production and rising consumer demand resulted in record-high global poultry imports\cite{PoultryEggsSector}. The poultry industry of the United States takes the lead domestically and internationally, supported by advanced production structures, modern poultry genetics, abundant domestic feed resources, and strong consumer demand\cite{PoultryEggsSector}. Global poultry sales reached \$70.2 billion in 2024, up 4\% from \$67.4 billion in 2023.

Despite its economic success, the poultry industry faces significant challenges during the processing phase, particularly within slaughter facilities. The working conditions in chicken processing plants are harsh because of low environmental temperatures around $4\,^{\circ}\mathrm{C}$ and high humidity\cite{demedeirosesperRobotisationIntelligentSystems2021}. In addition, workers are often required to work at rates dictated by 140 birds per minute processing line speeds, manually hanging chilled chicken carcasses on shackles for subsequent automated deboning and packaging tasks\cite{renAgriculturalRoboticsResearch2020a}. These repetitive and manual tasks not only result in a high incidence of workplace injuries but have also lead to a shortage of skilled labor\cite{diasEffectDifferentWorkrest2021}, which has forced the industry to seek inventive solutions that can mitigate worker safety concerns, maintain production efficiency, and ensure product quality.
\begin{figure}
    \centering
    \includegraphics[width=0.9\linewidth, angle=0]{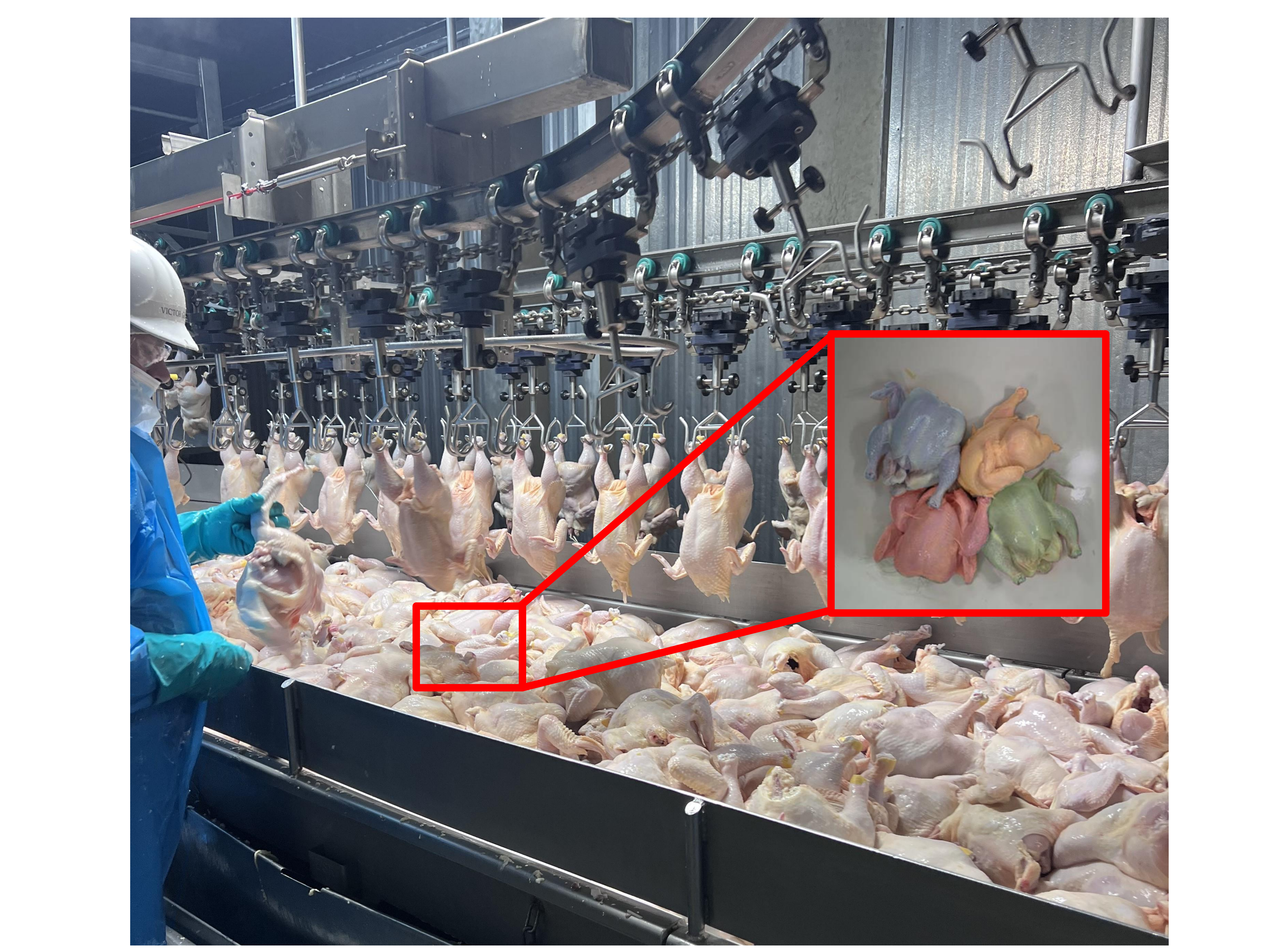}
    \caption{Chicken carcasses on a processing line in a poultry slaughter plant. The inset displays examples from our dataset with instances of overlapping carcasses overlaid with segmentation masks, where each color represents an individual chicken carcass.}
    \label{fig:enter-label}
\end{figure}

In recent years, the integration of automation technologies has offered great potential to improve poultry processing practices by reducing human exposure to hazardous conditions and improving production efficiency and food safety. In the poultry production phase, automation and robotics have made great progress in precision animal management\cite{pereira2020environmental}, such as monitoring environmental conditions, health management\cite{tong2024edge}, and egg picking\cite{renAgriculturalRoboticsResearch2020a}. However, in comparison, research on automation in poultry slaughter plants, especially for carcass handling is still limited \cite{yangDevelopmentTrendsChicken2024} because of unique challenges in image understanding\cite{sohrabipour2025cost} and robotic manipulation\cite{davar2025chicgrasp}. Challenges to image understanding, include (a) the variability in posture and shape of chicken carcasses, piling carcasses on conveyor belts where they exhibit diverse postures such as extended wings or bowed legs; (b) visual feature similarity, since all the carcasses are often stacked, and share similar colors and textures that makes visual differentiation of individual carcasses difficult; (c) occlusion and overlap, where carcasses are partially or fully obstructing one another on the conveyor; (d) the variations of lighting conditions, including shadows and reflections degrade image quality; Without intelligent visual understanding, robotic systems would struggle to manipulate chicken carcasses with the level of precision and accuracy necessary to meet industry standards.\cite{chowdhuryApplicationOpticalTechnologies2020}. This study focuses on the image understanding component and aims to improve image recognition accuracy and system robustness in these complex environments.

\section{RELATED WORK}
The integration of machine vision technologies has significantly advanced automation systems by providing intelligent input capabilities. Image instance segmentation and object detection are two main tasks in vision based industrial automation. 
 
For object detection, YOLO(you only look once)-based models have gained popularity for real-time object detection due to their speed and robustness in throughout the agri-food industry. In viticulture, YOLOv5s with a tracking algorithm enabled real-time grape cluster counting at 50.4 FPS with 84.9\% accuracy\cite{shen2023real}. For robotic weeding, a Multimodule-YOLOv7-L model achieved 97.1\% mAP at 37.3 FPS for lettuce identification and weed severity classification\cite{hu2024real}. YOLO models have also been effectively applied to detect apples, citrus, and other fruits under diverse conditions\cite{liu2024faster}. In poultry farming, YOLOv8x-DB supports dust bathing behavior detection in cage-free hens\cite{paneru2024tracking}, and the lightweight YOLO-Claw network achieved 97.1\% mAP for limb-health assessment\cite{wu2024yolo}, enabling precise welfare monitoring and health assessment for growing broilers.

While object detection models like YOLO based models are good at rapidly identifying and locating objects, many agricultural and food processing applications demand a more granular understanding, requiring not only detection but also precise pixel-level delineation. This need for detailed spatial information brings instance segmentation models to the forefront.

Instance segmentation models, such as Mask R-CNN and its variations, have been widely used in agriculture for tasks such as identifying and counting fruit, monitoring plant health, and detecting pests. These tasks require not only the detection and location of objects but also being able to differentiate among individual objects of the same class. This pixel-level classification is crucial in complex environments involving occlusions, variable shapes, and overlapping objects. Apples and strawberries are common benchmark crops in research due to their visual complexity and commercial value. For instance, a customized Mask R-CNN with deformable convolution and attention modules achieved a segmentation for apples under occlusion and varying lighting conditions\cite{wangFusionMaskRCNN2022}. Beyond the field, tasks often require objects handling and quality control where simple detection falls short. For instance, Mask R-CNN has been effectively applied to food item identification under diverse lighting conditions, showcasing its adaptability to visual challenges inherent in processing environments\cite{li2020enhanced}\cite{mohanty2022food}. More broadly, instance segmentation supports automated quality control, such as identifying defects in table olives\cite{macias2023mask}, and facilitates robotic bin-picking of soft, deformable food items like chicken fillets\cite{jonkerRoboticBinPickingPipeline2023} by providing precise location and shape information, even when items are overlapping. These applications show how instance segmentation's ability to deliver detailed, pixel-level understanding is critical for advancing automation and precision in food processing. 

Despite the effectiveness of deep learning-based visual models, their deployment and performance depends largely on large-scale annotated datasets, which are particularly difficult to obtain pixel-level annotation in poultry processing due to the complex visual features of carcasses, such as irregular shapes, overlapping limbs, soft, deformable tissue, and varying postures. Furthermore, variable lighting, reflections from wet surfaces, and contamination from feathers or blood exacerbate the difficulty of consistent object recognition \cite{attriReviewDeepLearning2023}. The Fig. 1 illustrates a representative example of these challenges which shows a typical overhead image from a poultry processing line, where multiple carcasses are partially overlapping, with irregular poses and shadow effects. Even for human annotators, precisely outlining each body part for segmentation is labor-intensive and error-prone. These issues highlight why robust instance segmentation is difficult to achieve without advanced data strategies.

To overcome the challenges of data collection and annotation, there are different image augmentation strategies that can help improve model robustness as well as resolve overlapping target objects by exposing individual objects to a variety of orientations and scales. Traditional image augmentation methods, such as flipping, rotating, cropping, and scaling, effectively expand the size of the training dataset\cite{lewyOverviewMixingAugmentation2023}. However, these affine transformation techniques have limited capabilities to introduce new, diverse instances of objects and are not able to simulate significant changes in object appearance, such as changes in texture, occlusion, or complex deformations\cite{huttenrauchLimitationsDataAugmentation2016}, \cite{zophLearningDataAugmentation2020}. These limitations are especially evident in poultry processing, where real-world unpredictability is notable, and traditional augmentations lack the ability to capture complex carcass handling conditions. 

Synthetic data generation provides a new and more advanced method to expand the dataset for neural network training. There are several approaches to generating the synthetic data. 3D modeling and rendering: the methods rely on physically-based rendering (PBR) engines to create synthetic scenes from scratch, simulating lighting, textures, and camera perspectives\cite{nikolenkoSyntheticDataDeep2021,gaur2023whale}. 
Recent advances have shifted toward neural rendering methods such as Neural Radiance Fields (NeRF), which model volumetric scenes from multiple views to synthesize realistic novel perspectives\cite{mildenhall2021nerf}, and 3D Gaussian Splatting (3DGS), which represents surfaces using anisotropic Gaussian primitives for efficient real-time rendering\cite{kerbl2023gaussian}. Additional applications encompass strawberry plant disease detection systems, where synthetic imagery facilitates model training for identifying various pathological stages under diverse environmental conditions\cite{aghamohammadesmaeilketabforooshEnhancingStrawberryDisease2024}. These neural methods enable a more faithful replication of object-level details, dynamic geometry, and realistic occlusion. By leveraging these methods, researchers can generate complex synthetic datasets that more accurately reflect real-world variability and lighting conditions. 

Building upon these advanced visual rendering techniques, physics-based simulation imitates real-world dynamics\cite{karMetasimLearningGenerate2019}, which offers another pathway for synthetic data generation, focusing on reproducing the dynamic behaviors and interactions among objects and environments beyond just their visual appearance\cite{todorov2012mujoco}. This approach generates data by emulating real-world physical laws, ensuring that the synthetic contents are not only visually plausible but also that their behavior, such as object motion, collisions, deformations, fluid dynamics, and environmental changes all adhere to physical logic. Consequently, researchers can create large datasets encompassing diverse complex scenarios, edge cases, or interactions that would be difficult to replicate in the the actual processing plant\cite{lin2025exploring}\cite{gan2020threedworld}, thereby enhancing the robustness and generalization capabilities of machine learning models.

The combination of these two approaches is particularly powerful: physics-based simulation can generate 3D scenes and events with complex dynamic behaviors and physical interactions, while neural rendering techniques like 3DGS can efficiently render these dynamic, physically plausible scenes from arbitrary new viewpoints with high fidelity. This synergy allows the generated synthetic data to achieve high standards in both visual realism and behavioral authenticity, providing invaluable data resources for training more robust vision models.

In addition, style transfer and image synthesis using generative models can produce new data variations for training purposes\cite{wangHighresolutionImageSynthesis2018}. Generative Adversarial Network (GAN)\cite{goodfellow2020generative} is one of the generative frameworks designed to synthesize new data with the same characteristics that visually resemble the data in the training set\cite{goodfellowGenerativeAdversarialNets2014,karabatis2023detecting}. In agriculture community, GANs enhance dataset have been used for tasks like weed segmentation \cite{tianDetectionAppleLesions2019}, apple detection\cite{fawakherjiShapeStyleGANbased2024}, and oyster recognition\cite{lin2023oysternet}. GANs improve model accuracy and robustness, addressing challenges like data scarcity and environmental variability. However, GANs have limitations, including mode collapse, where the model generates limited diversity in outputs, and training instability due to the adversarial process. Additionally, GAN-generated data may introduce biases, potentially affecting model accuracy in real-world applications\cite{wangLycheeSurfaceDefect2021}; 

Another prominent generative framework is the diffusion model\cite{podell2023sdxl}\cite{ho2020denoising}. These models operate through a two-stage process: systematically adding noise to data and then training a neural network to remove the noise, iteratively by transforming the added noise into coherent data. Diffusion models have gained acceptance for their ability to generate high-fidelity and diverse images\cite{dhariwal2021diffusion}, often surpassing other models in sample quality and training stability. This makes diffusion models promising for specialized applications, such as generating synthetic oysters for aquaculture\cite{lin2024odyssee,campbell2025ai} or augmenting datasets for rare plant diseases or produce defects. However, a primary limitation of these models is their slow, iterative sampling process, which can be computationally intensive compared to single-pass generation methods\cite{song2020denoising}. Continuing to address these sampling speed challenges and explore their utility in various scientific domains. 

Although most synthetic data generation methods significantly increase the quantity and visual realism of training datasets, they still do not address one of the most persistent difficulties in deep learning: the time consuming and labor intensive task of manual annotation. Even when synthetic images are available, many applications still require manual masks or bounding boxes annotation to ensure model accuracy. This makes it difficult to scale up datasets efficiently. 

In this paper, we propose a practical framework that generates automatically labeled synthetic data and combines it with real datasets under various configurations to enhance instance segmentation performance using an initial example from the poultry processing industry. We evaluated three leading deep learning models: Mask R-CNN, Mask2Former, and YOLOv11 across five dataset settings, including a real chicken carcass dataset baseline and four augmented datasets with up to 1000 Blender-generated synthetic images. Blender, a 3D rendering platform allows for precise control of the scene parameters and automatic generation of accurate annotations, offers greater structural fidelity and annotation reliability compared to other generative models like GANs or diffusion methods, which are particularly important in tasks involving non-rigid, anatomically varied, and piled objects. Our study focuses on the rehanging phase of poultry processing,where entact broiler carcasses are typically piled on a take-away belt when exiting the cold-water chiller and must be rehung on an overhead, moving schackle line. It is critical that reliable real-time segmentation will be available for the next phase, using robots to perform this precise identification of the orientation of the piled poultry carcass, pick-up an individual carcass and hang it by the hocks on the moving schackle line. This study presents a simulation-to-reality (Sim2Real) framework for generating synthetic training data targeting the segmentation of occluded chicken carcasses in dense configurations. This work supports broader efforts to improve the worker safety, efficiency, consistency, and scalability of poultry production systems through data efficient solutions.

\section{METHODOLOGY}
This study elaborates on the collection of real-world and synthetic chicken carcass segmentation datasets, subsequently proposing a dataset augmentation method for improving poultry instance segmentation performances. In this study, a deep neural network was trained on a combination of a small-scale annotated real-world data and large-scale synthetic data rendered from Blender (a computer graphics software).  In this experiment, the real-world data was collected using a custom hardware system, as shown in Fig. 2, however, obtaining such datasets with annotation requires significant time and resources. To address this challenge, this study adopted a ``Sim-to-Real" strategy, which depends on a task-specific and data-efficient solution. By using Blender to simulate anatomically plausible carcass identification and positions and generate high-fidelity ground-truth masks, our framework directly addresses key obstacles in this specific dataset: annotation bottlenecks and visual complexity. A comparison analysis of competition deep learning models provide strong support for this method's ability to improve segmentation accuracy and enhance practical feasibility.

\subsection{Real-world dataset collection}
We built a real-world poultry dataset using a custom image acquisition system designed to capture diverse scenarios. To ensure the sample consistency, all poultry carcasses were obtained from local grocery stores affiliated with the same supplier network that follows standardized packaging and cold-chain procedures, which helped reduce variation in carcass appearance. To simulate various real-world conditions, poultry carcasses were randomly arranged in different positional combinations, including stacked and closely placed configurations. 
\begin{figure}
    \centering
    \includegraphics[width=0.9\linewidth, trim=800 400 750 600, clip,angle=0]{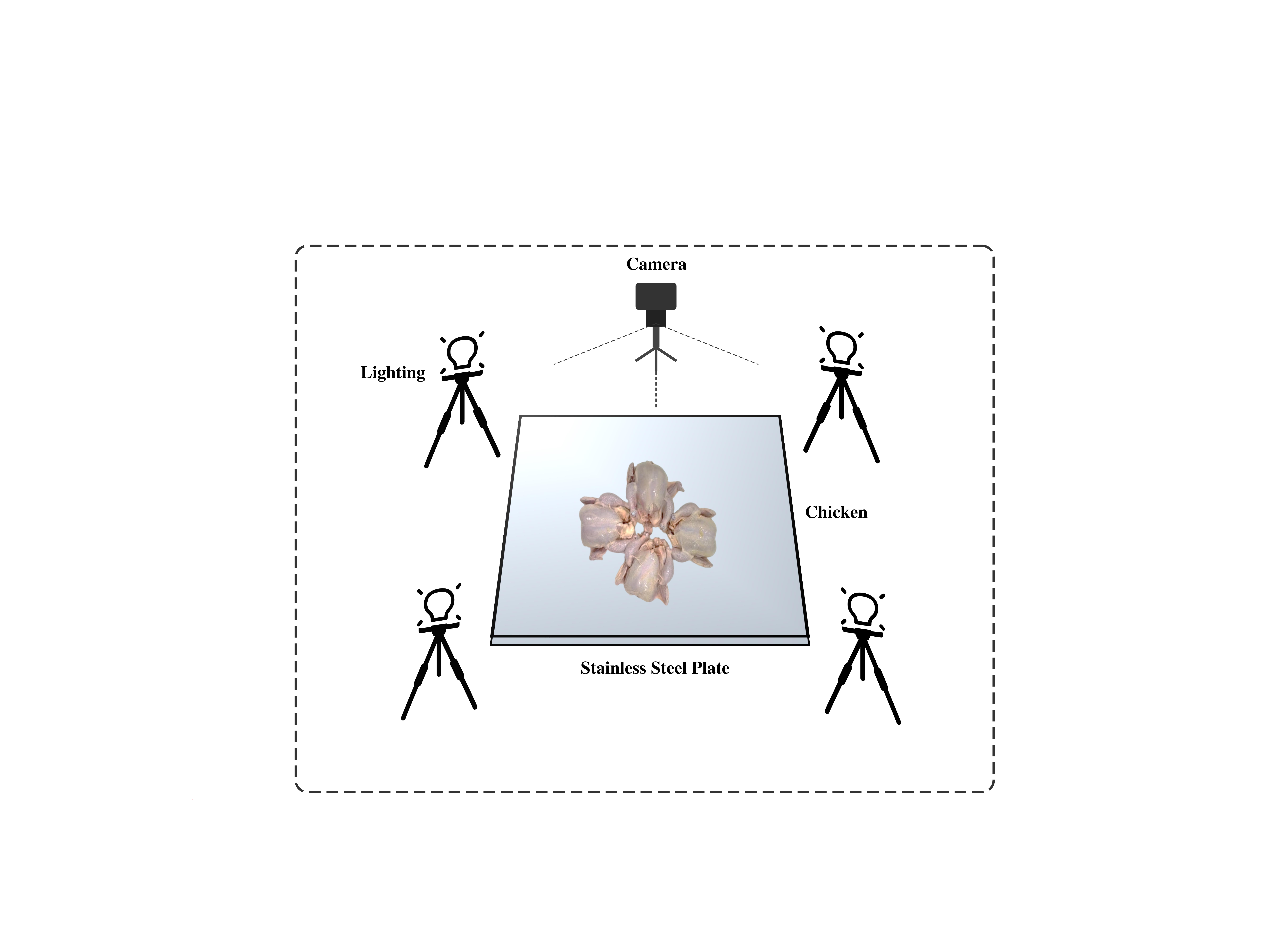}
    \caption{The real chicken data acquisition system}
    \label{fig:data acquisition system}
\end{figure}
The image acquisition system was equipped with controlled lighting to maintain consistent illumination throughout the image capture process, as shown in Figure 2. The background of the poultry carcasses was a stainless-steel plate, mimicking the stainless steel food contact surfaces used in poultry production. In total, 300 real world images were collected, which were composed of images with different carcass overlays and image capturing angles. These images form the baseline of our dataset, providing a robust performance evaluation of our segmentation model under varying conditions. Each image was manually annotated with pixel-level instance segmentation masks using LabelMe\cite{russell2008labelme}. This collection of 300 annotated real-world images was then divided into distinct training, validation, and testing sets to develop and evaluate our segmentation model. Additional details are discussed in Section C. 

\subsection{Synthetic Dataset Generation}
Blender is an open-source 3D creation software, has extensive modeling, rendering, and animation features\cite{denningerBlenderprocReducingReality2020}. By using tools like Blender, researchers can generate synthetic datasets that closely mimic real-world scenarios\cite{nikolenkoSyntheticDataDeep2021}\cite{lin2023oysternet}\cite{gaur2023whale}.  Its Python API can be used to automatically generate synthetic datasets by altering the camera positions, lighting conditions, and object properties. Ground truth labels are automatically generated during the rendering process, which allows the researchers to quickly build large-scale synthetic datasets with accurate annotations. The software's physically-based rendering engine can simulate realistic lighting, shadows, and material properties, which making the synthetic images more representative of real-world conditions.

\begin{figure}
    \centering
    \includegraphics[width=\linewidth,trim=800 500 1000 400, clip, angle=0]{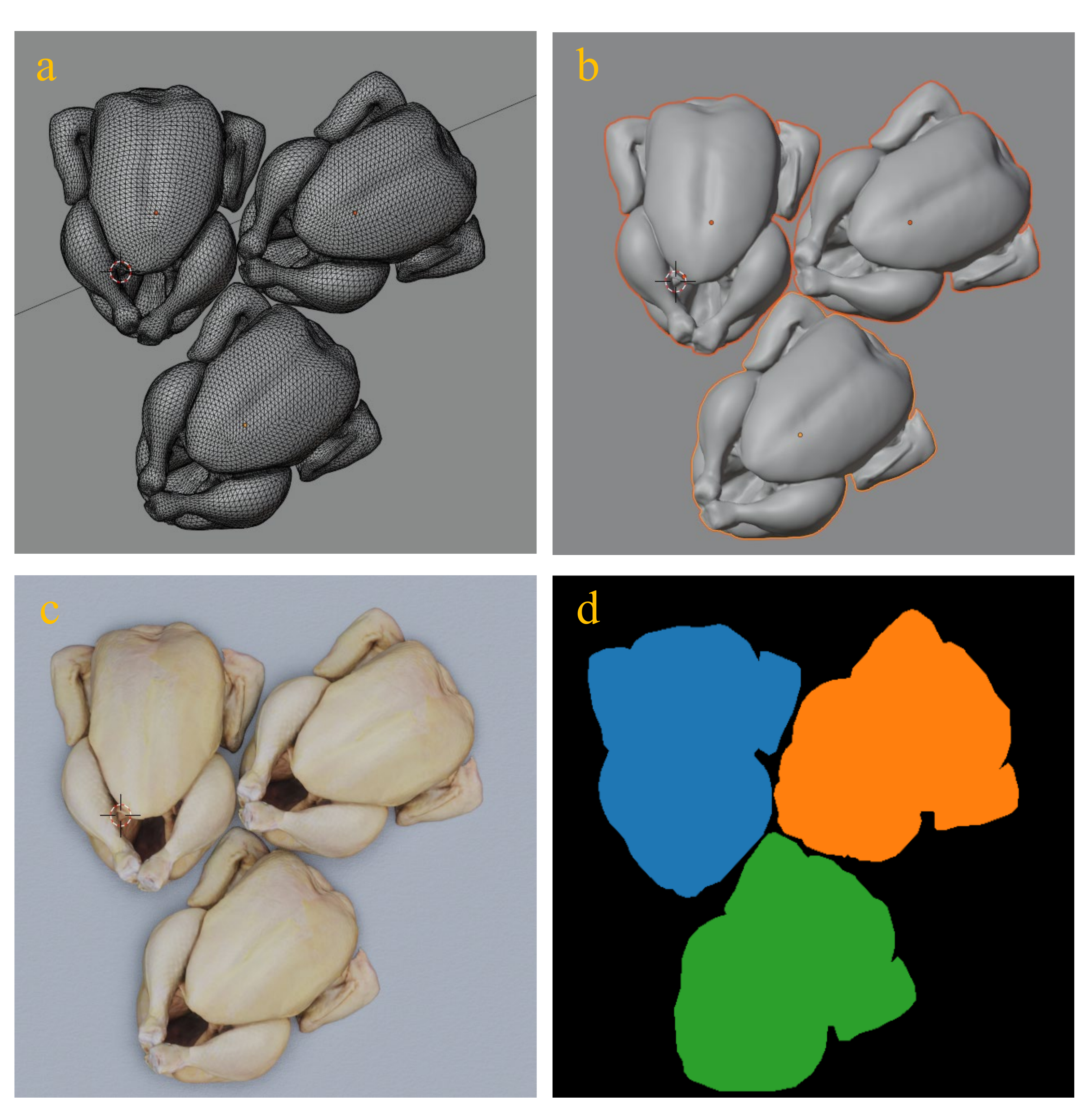}
    \caption{An overview of our synthetic data generation workflow. (a) Mesh Preparation: High quality 3D chicken models are imported into Blender and arranged in a non-overlapping layout. Mesh resolution and topology are inspected in wireframe mode. (b) Geometry Refinement: Surface smoothness, orientation, and model boundaries are adjusted under solid viewport shading to ensure clean geometry before rendering. (c) RGB Rendering: Realistic lighting, textures, and materials are applied to generate photorealistic synthetic images using the Cycles engine. (d) Mask Generation: Each instance is assigned a unique ID to generate the corresponding segmentation masks. These automated generated masks are used as ground truth for training instance segmentation models.}
    \label{fig:synthetic data workflow}
\end{figure}

Blender has demonstrated considerable efficiency across diverse agricultural applications for synthetic data generation. Barth et al. utilized Blender to generate a comprehensive dataset comprising 10,500 synthetic images of Capsicum annuum through systematic randomization of 3D plant models based on empirical morphological measurements. This dataset was specifically designed to replicate visual conditions prevalent in greenhouse environments and facilitate instance segmentation tasks in precision agriculture \cite{barthDataSynthesisMethods2018}. 
The ability of Blender-generated synthetic data to substantially enhance model performance in food processing applications, by effectively capturing complex real-world variations, has been highlighted by several studies. For instance, Ummadisingu et al. demonstrated this enhancement for food item segmentation in meal-assisting robotic systems\cite{ummadisinguClutteredFoodGrasping2022}, and Jonker et al. showcased its benefits for chicken fillet handling in automated robotic systems\cite{jonkerRoboticBinPickingPipeline2023}.

While previous applications have demonstrated Blender's capabilities in agricultural contexts, poultry processing presents unique challenges that have been largely unaddressed. Unlike the aforementioned applications involving static plant models or food items, poultry carcass handling has complex anatomical variations from each carcass. Our approach leverages Blender's capabilities to develop comprehensive training datasets that capture the intricate variations inherent in industrial poultry processing, including realistic simulation of diverse carcass orientations and anatomical overlap typically encountered during rehanging operations\cite{sohrabipour2025cost}. This represents a significant advancement beyond existing applications by addressing the dynamic and morphologically complex nature of biological products typically encountered in food processing environments.

\begin{figure*}[htbp]
    \centering
    \includegraphics[width=\linewidth, trim=100 550 100 400, clip,angle=0]{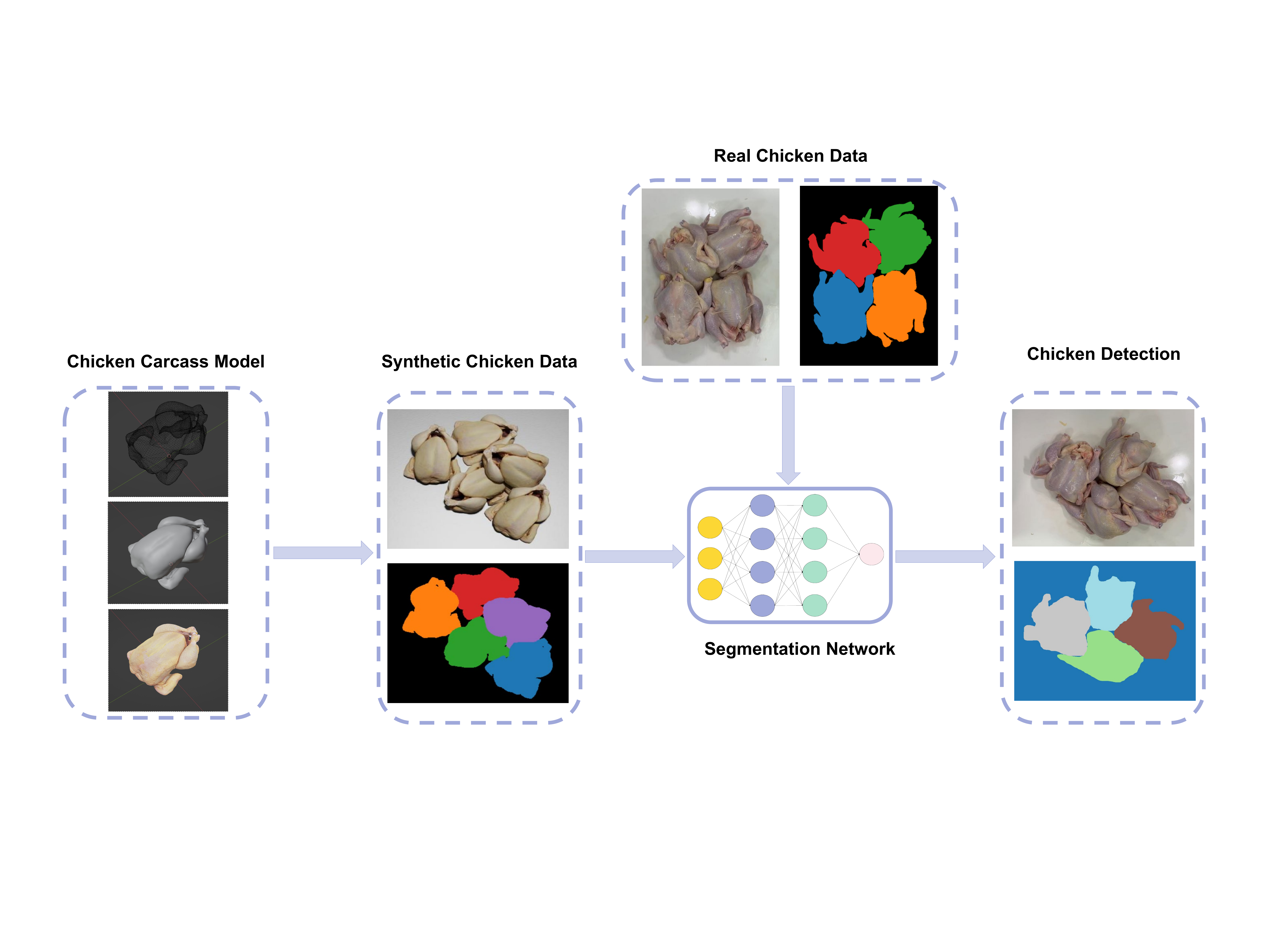}
    \caption{An overview for Synthetic data generation and model training. The process begins with a high fidelity 3D chicken carcass model, which is used to generate a large-scale synthetic chicken dataset. This synthetic data is then combined in varying quantities with our real-world training data to form a series of distinct hybrid datasets. Each datasets are subsequently used to train and evaluate the deep learning Segmentation Networks (including Mask R-CNN with ResNet-50/101 backbones, Mask2Former, and YOLOv11-seg). This comprehensive approach allows for a systematic investigation of how synthetic data impacts performance across different art architectures. The trained network can then perform robust instance segmentation (Chicken Detection) on new, unseen real-world images, accurately identifying and delineating each individual carcass even in cluttered scenes. }
    \label{fig: Segmentation overview}
\end{figure*}

In poultry processing plants, chicken carcasses are typically spread or stacked on flat surfaces during inspection and handling procedures. Our study focuses on addressing the challenges specific to poultry carcass instance segmentation, such as overlapping and close positioning, which make it difficult for visual models to distinguish among individual carcasses. These complex spatial arrangements also present significant challenges for human annotators, making manual labeling both time consuming and error prone, which highlights the advantages of the Blender approach. 

In the virtual simulation, carcasses were placed in configurations ranging from isolated to tightly clustered or overlapped. As shown in Fig. \ref{fig:synthetic data workflow}, each carcass was treated as an individual input in the mask annotations, while the background remained uniform to avoid unnecessary complexity. The top-view perspective was chosen to reflect practical application scenarios of how the human operator sees the carcasses in production environments. Using this approach, a total of 1000 synthetic images were generated, each containing RGB images and the corresponding instance masks with COCO format. The synthetic dataset provided a robust training data, ensuring the model learns to handle overlapping and closely positioned carcasses effectively. 

\subsection{Deep Neural Network and training details }
This paper implemented Mask R-CNN\cite{he2017mask}, Mask2Former\cite{cheng2022masked}, and YOLOv11-seg\cite{wang2023yolov11} as the baseline instance segmentation models for our analysis. Each model was systematically evaluated to determine optimal performance characteristics for industrial poultry processing applications. Followings are some implementation details:

For Mask R-CNN, the model was deployed with two different backbone configurations (ResNet-50-FPN and ResNet-101-FPN) to evaluate the impact of network depth on segmentation performance. Both backbones integrated Feature Pyramid Network (FPN) to hierarchically combine multi-scale features, enhancing semantic representation across resolution levels. The Region Proposal Network (RPN) generated candidate object regions through sliding window analysis, proposing anchors that underwent refinement to localize potential carcass instances. These refined proposals subsequently facilitated the prediction head's triple task of classification, bounding box regression, and mask generation, yielding precise per-instance segmentation masks for poultry carcasses. For Mask2Former,  the model was implemented with a Swin Transformer Tiny (Swin-T) backbone, which offers a hierarchical representation with shifted windows for efficient self-attention computation. This model replaced the traditional convolutional backbone with a transformer architecture, providing enhanced capability to capture global context while maintaining computational efficiency. For the YOLO model, the latest YOLO architecture (YOLOv11) was utilized with its segmentation capability, which offered a one-stage detection approach with improved speed-accuracy trade-offs compared to traditional two-stage detectors.

\begin{table}[htbp]
\centering
\caption{Chicken detection models trained with real and synthetic datasets.}
\label{tab:datasets}
\begin{tabular}{llcccccc}
\toprule[1.5pt]
\textbf{Setting} & \textbf{Real Train} & \textbf{Real Val} & \textbf{Real Test} & \textbf{Synthetic Train} \\
\midrule
Real\_Baseline   & 60  & 60  & 180 & 0     \\
\midrule
Rea+Syn-250     & 60  & 60  & 180 & 250   \\
\midrule
Rea+Syn-500     & 60  & 60  & 180 & 500   \\
\midrule
Rea+Syn-750     & 60  & 60  & 180 & 750   \\
\midrule
Rea+Syn-1000    & 60  & 60  & 180 & 1000  \\
\bottomrule[1.5pt]
\end{tabular}
\end{table}

All models were trained with consistent training hyperparameters to ensure fair comparisons: 200 epochs for all models, batch sizes of 8, image resolution of 640×640 pixels, utilizing NVIDIA A100 GPU (40GB memory) provided by the Arkansas High Performance Computing Center (AHPCC), and data augmentation including random flipping (probability=0.5), random cropping (512×512), and normalization. For Mask R-CNN and Mask2Former, a step-wise learning rate schedule with linear warmup over 500 iterations and learning rate decay at epochs 30, 60, and 100. Mask R-CNN was optimized using SGD with momentum (0.9) and weight decay (0.0001), while Mask2Former utilized AdamW optimizer with parameter-wise learning rates to account for the different components in the transformer architecture\cite{loshchilov2017decoupled}\cite{cheng2022masked}. For YOLOv11, the default optimizer configuration is with an initial learning rate of 0.01, momentum of 0.937, and weight decay of 0.0005. 

The experimental design incorporated a systematic approach to evaluate the impact of synthetic data augmentation on model performance. The 300 real-world images were randomly partitioned into training (20\%), validation (20\%), and test (60\%) sets, with a fixed random seed to ensure reproducibility. We allocated less data in training set and more data in the test set to prove the effectiveness of using the synthetic data for model trianing. As shown in Table \ref{tab:datasets}, to evaluate the contribution of synthetic data, five training combinations of the real images and synthetic data. The baseline models were trained solely on 60 real-world images, and four additional settings will include the same number of real-world data supplemented by 250, 500, 750, and 1000 synthetic images, respectively. The validation (60 images) and test (180 images) datasets remain unchanged which comprised exclusively of real-world poultry processing imagery, ensuring fair assessment of how synthetic data augmentation affects model generalization to actual industrial conditions.

\subsection{Evaluation Metrics}
To evaluate the performance of the poultry carcass segmentation model, the COCO dataset evaluation plugin provided in PyTorch, which is a standard for instance segmentation tasks. Each model processed RGB imagery and generated instance predictions comprising segmentation masks, confidence values, and localization boundaries for individual poultry carcasses. Performance evaluation centered on the instance segmentation quality using mask overlap analysis. We utilized the spatial coincidence ratio between predictions and ground-truth annotations, calculated as:

\begin{equation}
\mathrm{IoU} = \frac{\text{Area of overlap}}{\text{Area of union}}.
\end{equation}

The primary metric used for assessment was mean Average Precision (mAP), which provides a comprehensive measure of segmentation accuracy. The analysis framework reported three principal metric variants: AP\textsubscript{50} (average precision at IoU threshold 0.5), AP\textsubscript{75} (at IoU threshold 0.75), and AP (mean average precision across all IoU thresholds from 0.5 to 0.95). These metrics were calculated separately for both instance masks and bounding boxes to evaluate segmentation and localization performance independently.

\section{RESULTS}
Table~\ref{tab:model-results} summarises the quantitative performance of four
instance-segmentation frameworks, MaskR-CNN with ResNet-50 (R50) and ResNet-101 (R101) backbones, Mask2Former, and YOLOv11-Seg trained on five different dataset settings. The real-only configuration is denoted \textbf{Real\_Baseline}, whereas \textbf{Real+Syn-$N$} ($N\!\in\!\{250,500,750,1000\}$) indicates synthetic images were added to the same real set.  All models were evaluated on same 180-real world images in the test set using the COCO-style mAP, which averaged precision over IoU thresholds from 0.50 to 0.95. AP\textsubscript{50} and AP\textsubscript{75} were reported to capture low and high IoU performance.

\begin{table*}[htbp]
\centering
\caption{Performance of Instance Segmentation Models Trained with Varying Amounts of Synthetic Data.}
\label{tab:model-results}
\begin{tabular}{llcccccc}
\toprule[1.5pt]
\textbf{Model} & \textbf{Setting} & \textbf{bbox\_mAP} & \textbf{bbox\_mAP\_50} & 
\textbf{bbox\_mAP\_75} & \textbf{segm\_mAP} & \textbf{segm\_mAP\_50} & \textbf{segm\_mAP\_75} \\
\midrule
\multirow{5}{*}{Mask R-CNN (R50)} 
  & Real\_Baseline & 0.740 & 0.963 & 0.866 & 0.746 & 0.965 & 0.863 \\
  & Real+Syn-250   & 0.761 & 0.962 & 0.861 & 0.736 & 0.954 & 0.838 \\
  & Real+Syn-500   & 0.784 & 0.959 & 0.883 & 0.755 & 0.959 & 0.865 \\
  & Real+Syn-750   & 0.795 & 0.962 & 0.896 & 0.765 & 0.961 & 0.877 \\
  & Real+Syn-1000  & 0.793 & 0.960 & 0.886 & 0.759 & 0.950 & 0.871 \\
\midrule
\multirow{5}{*}{Mask R-CNN (R101)} 
  & Real\_Baseline & 0.750 & 0.960 & 0.869 & 0.733 & 0.956 & 0.836 \\
  & Real+Syn-250   & 0.795 & 0.962 & 0.883 & 0.751 & 0.952 & 0.856 \\
  & Real+Syn-500   & 0.773 & 0.961 & 0.884 & 0.751 & 0.951 & 0.857 \\
  & Real+Syn-750   & 0.776 & 0.964 & 0.874 & 0.753 & 0.963 & 0.860 \\
  & Real+Syn-1000  & 0.805 & 0.962 & 0.881 & 0.766 & 0.953 & 0.875 \\
\midrule
\multirow{5}{*}{Mask2Former} 
  & Real\_Baseline & 0.434 & 0.712 & 0.437 & 0.523 & 0.795 & 0.597 \\
  & Real+Syn-250   & 0.530 & 0.776 & 0.548 & 0.669 & 0.899 & 0.739 \\
  & Real+Syn-500   & 0.652 & 0.896 & 0.694 & 0.737 & 0.933 & 0.835 \\
  & Real+Syn-750   & 0.672 & 0.896 & 0.715 & 0.726 & 0.925 & 0.809 \\
  & Real+Syn-1000  & 0.661 & 0.880 & 0.694 & 0.737 & 0.927 & 0.816 \\
\midrule
\multirow{5}{*}{YOLOv11-seg} 
  & Real\_Baseline & 0.835 & 0.969 & 0.890   & 0.784 & 0.959 & 0.846 \\
  & Real+Syn-250   & 0.845 & 0.965 & 0.891   & 0.817 & 0.958 & 0.874 \\
  & Real+Syn-500   & 0.842 & 0.966 & 0.888   & 0.817 & 0.964 & 0.883 \\
  & Real+Syn-750   & \textbf{0.857} & 0.966 & \textbf {0.906}   & 0.828 & 0.965 & 0.891 \\
  & Real+Syn-1000  & 0.852 & \textbf{0.972} & 0.896   & \textbf{0.833} & \textbf{0.971} & \textbf{0.898} \\
\bottomrule[1.5pt]
\end{tabular}
\end{table*}
\subsection{Overall trends}

For all models, adding synthetic images consistently improved both detection and segmentation metrics. However, the extent of this improvement and the optimal amount of synthetic data depended on the model architecture and the specific evaluation metric. For example, the Mask2Former model, which had a lower baseline, showed a substantial relative increase with synthetic data. More robust baseline models like YOLOv11-seg also benefited, enhancing the high performance more, particularly with larger volumes of synthetic images. There was not always a linear improvement with the addition of more synthetic data; for some models, a diminishing return or even a slight decrease in performance was observed after a certain amount of synthetic data was added. For instance, Mask R-CNN R50 beyond 750 synthetic images.

\begin{figure*}[htbp]
    \centering
    \includegraphics[width=\linewidth, trim=80 520 100 690, clip,angle=0]{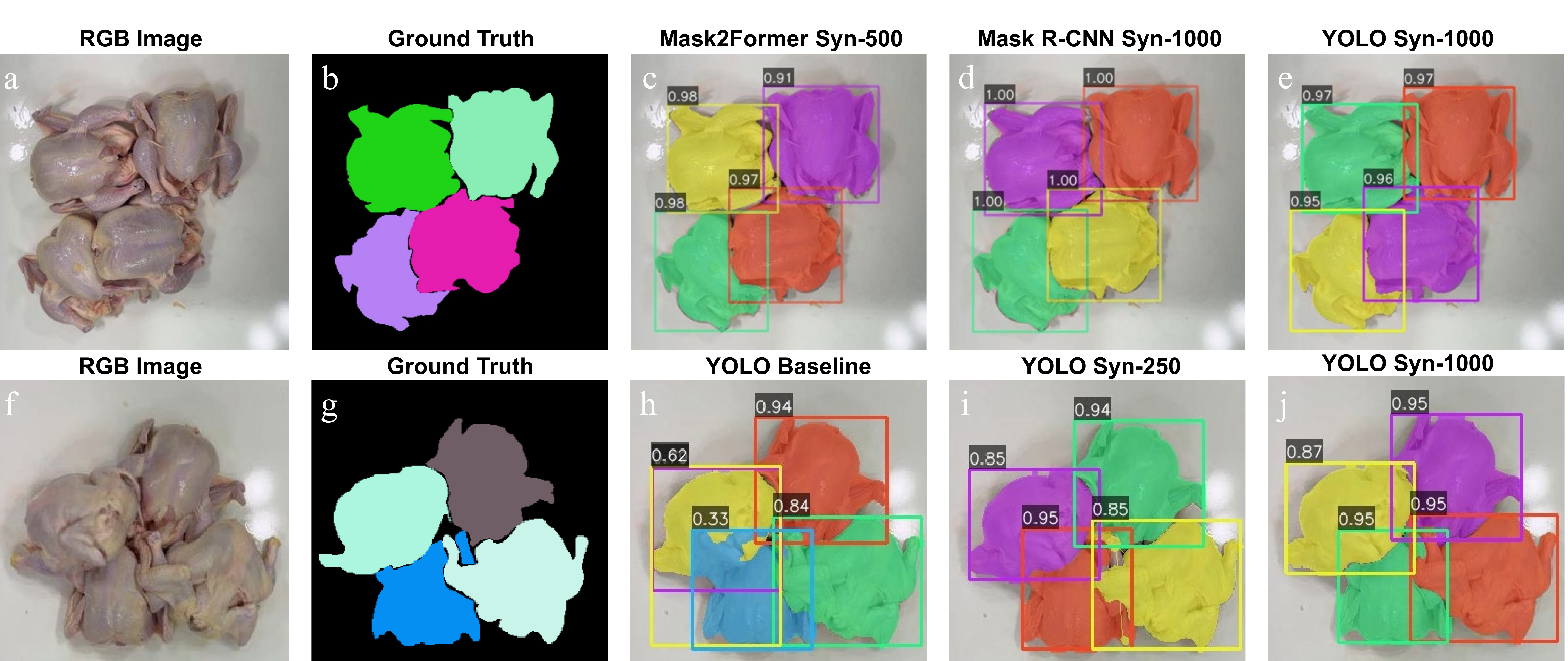}
    \caption{Comparison of bounding box and instance segmentation results obtained with the best-performing training ratios for each model (Mask-RCNN with ResNet-101 backbones, Mask2Former, and YOLOv11-seg) and with progressively larger synthetic sets for YOLOv11-seg. Example predictions on two real chicken-carcass images.
    (a, f) Input RGB images.
    (b, g) Ground-truth instance masks.
    (c–e) Predictions for the raw image (a) using the three baseline architectures, each shown with the training configuration that yielded its highest COCO mAP;(0.50 : 0.95): (c) Mask2Former trained with Real + Syn-500, (d) Mask R-CNN (ResNet-101) trained with Real + Syn-1000, and (e) YOLOv11-seg trained with Real + Syn-1000.
    (h–j) Predictions for image (f) from YOLOv11-seg models trained with increasing synthetic-to-real ratios: (h) Real-only baseline, (i) Real + Syn-250, and (j) Real + Syn-1000. Masks are colour-coded per instance and overlaid with the associated bounding box and confidence score. }
    \label{fig: detection results}
\end{figure*}

\subsection{Effect of the synthetic to real ratio}

The real training dataset was set to 60 images. The experiments explored synthetic-to-real data ratios by adding 0, 250, 500, 750, or 1000 synthetic images, corresponding to ratios of 0:60 (baseline), approximately 4.17:1 (Syn-250), 8.33:1 (Syn-500), 12.5:1 (Syn-750), and 16.67:1 (Syn-1000).  For Mask R-CNN (R50), a ratio around 12.5:1 appeared optimal for both \textit{bbox\_mAP} (0.795) and \textit{segm\_mAP} (0.765). For Mask R-CNN (R101), the highest \textit{bbox\_mAP} (0.805) and \textit{segm\_mAP} (0.766) were achieved at the largest ratio of approximately 16.67:1, suggesting it could leverage more synthetic data effectively than its R50 model. Mask2Former showed impressive gains up to a ratio of 8.33:1, achieving \textit{bbox\_mAP} of 0.652 and \textit{segm\_mAP} of 0.737. Further increases to 12.5:1 yielded the peak \textit{bbox\_mAP} (0.672), while \textit{segm\_mAP} was similar or peaked at 16.67:1 with 0.737. YOLOv11-Seg consistently benefited from increasing synthetic-to-real ratios. The best \textit{bbox\_mAP} (0.8570) was at a 12.5:1 ratio, while the highest \textit{segm\_mAP} (0.8325) and several other key metrics (\textit{bbox\_mAP\_50}, \textit{segm\_mAP\_50}, \textit{segm\_mAP\_75}) peaked at the 16.67:1 ratio.

Those results indicate that different model capacities and architectures respond differently to the proportion of synthetic data. While lighter models like R50 stop improving significantly as more synthetic data is added, but deeper or transformer-based architectures continue to benefit from larger synthetic datasets.
\subsection{Model-Specific Performance Analysis}
\subsubsection{Mask R-CNN (R50 and R101) and Backbone Comparison}

\begin{itemize}
    \item Mask R-CNN (R50): The baseline $bbox\_mAP$ was 0.740 and $segm\_mAP$ was 0.746. Adding synthetic data generally improved performance, with the ``Real+Syn-750" setting yielding the best results ($bbox\_mAP$ 0.795, $segm\_mAP$ 0.765). Performance slightly decreased with 1000 synthetic images.
    \item Mask R-CNN (R101): The baseline $bbox\_mAP$ (0.750) was slightly higher than R50, but $segm\_mAP$ (0.733) was lower. The R101 backbone generally benefited from more synthetic data than R50, achieving its peak $bbox\_mAP$ (0.805) and $segm\_mAP$ (0.766) with 1000 synthetic images.
    \item Backbone Comparison (R50 vs. R101): The deeper R101 backbone, achieved a slightly higher peak $bbox\_mAP$ and $segm\_mAP$ (0.805 and 0.766 respectively with 1000 synthetic images) compared to the R50 backbone's peak (0.795 $bbox\_mAP$ and 0.765 $segm\_mAP$ with 750 synthetic images). The R101 also seemed to better utilize larger amounts of synthetic data (peaking at 1000 images) compared to R50 (peaking at 750 images). However, the baseline performance of R101 for segmentation was only slightly lower than R50's. At their respective optimal synthetic data settings, R101 slightly outperformed R50.
\end{itemize}
\subsection{Bounding Box Detection and Instance Segmentation Quality}

The qualitative impact of these metric improvements is visualized in Figure. \ref{fig: detection results}. The top row directly compares the best configurations of the three architectures. While all models perform well, notable differences in detail are apparent. For instance, focusing on the bottom-left carcass, the mask contour from YOLOv11-seg (e) provides the tightest fit to the ground-truth edges, whereas the masks from Mask R-CNN (R101) (d) and Mask2Former (c) exhibit a degree of under-segmentation in this area, failing to capture the full details of the object.

The bottom row of Figure. \ref{fig: detection results} provides a clear qualitative illustration of the value of synthetic data. While the baseline model (h) detected all instances, its segmentation quality and confidence scores were notably deficient, particularly for the two lower, touching carcasses. This performance is substantially improved with the addition of synthetic data. When increasing to 1000 synthetic images (j), the mask delineations for these two lower carcasses become significantly more precise, and their confidence scores dramatically increase from 0.33 and 0.84 in (h) to 0.95 for both in (j). This visually demonstrates that synthetic data not only boosts the model's detection confidence but also effectively refines segmentation boundaries in complex, overlapping scenes.

Generally, bounding box detection ($bbox\_mAP$) were similar with instance segmentation ($segm\_mAP$), but with some difference:
\begin{itemize}
    \item For Mask R-CNN (R50), the optimal amount of synthetic data (Syn-750) was the same for both $bbox\_mAP$ and $segm\_mAP$.
    \item For Mask R-CNN (R101), 1000 synthetic images yielded the best results for both $bbox\_mAP$ and $segm\_mAP$.
    \item For Mask2Former, $bbox\_mAP$ peaked with 750 images, while $segm\_mAP$ peaked with 500 and 1000 synthetic images, indicating a slightly different optimal point or a plateau for segmentation.
    \item For YOLOv11-seg, $bbox\_mAP$ peaked with 750 synthetic images, while $segm\_mAP$ (and related $segm\_mAP_{75}$) continued to improve up to 1000 synthetic images, suggesting that segmentation quality, which requires more precise localization, might benefit more from larger synthetic datasets with this model.
\end{itemize}

The $mAP_{50}$ scores (for both bbox and segm) were generally very high across all models and settings with synthetic data, often above 0.95 for Mask R-CNN and YOLOv11-seg. This indicated that all models became proficient at detecting and segmenting the easier instances (IoU threshold of 0.5). These improvements from synthetic data were often more pronounced in the stricter $mAP_{75}$ metric and the overall $mAP$ (0.50:0.95), highlighting that synthetic data helped improve the precise localization capabilities of the models. For example,  the $segm\_mAP_{75}$ of YOLOv11-seg increased 5.72\% from 0.8461 (baseline) to 0.8975 (with 1000 synthetic images).

\subsection*{Summary of Results} 

The addition of synthetic data was broadly beneficial for all instance segmentation models tested, improving both bounding box detection and instance segmentation performance. YOLOv11-seg demonstrated the highest overall performance, effectively leveraging up to 1000 synthetic images to achieve top scores, particularly in segmentation metrics. This quantitative superiority was also reflected in the qualitative analysis, where YOLOv11-seg consistently produced more precise segmentation masks compared to the other architectures. Mask R-CNN models also showed consistent improvements, with the R101 backbone benefiting from a larger volume of synthetic data than the R50. Mask2Former, while starting from a lower baseline, exhibited significant relative gains, underscoring the value of synthetic data for this architecture. Notably, for several models, an optimal balance of synthetic to real data was identified, beyond which performance gains plateaued or, in some isolated cases, slightly decreased. The ideal amount of synthetic data and the resulting performance gains depended on the models, and these improvements were typically more pronounced at higher IoU thresholds. The visual evidence strongly supported this, qualitatively demonstrating how synthetic data transformed low-confidence, imprecise detections into high-confidence, accurate segmentations, especially in challenging cases of object overlap.

\section{Discussion}

In this paper, the observed performance improvements across various instance segmentation models when augmenting real training data with synthetic images can be attributed to several factors, which align with, and in some cases nuance, findings from previously published literature. Primarily, the introduction of synthetic data likely increased the diversity of training samples. This exposed the models to a wider range of chicken carcass orientations, occlusions, and background variations than were available in our limited real dataset (60 images). Such enhanced diversity generally helps models generalize better to unseen data and reduces overfitting to the specific characteristics of a small real dataset. For instance, synthetic data can provide numerous examples of challenging scenarios, such as chicken carcasses in dense clusters or under unusual lighting, which might be rare or absent in limited real-world imagery. This principle is well-supported by other studies. For example, early work by Richter et al. demonstrated the enhanced performance of using synthetic data from game engines for semantic segmentation\cite{richter2016playing}, and Tremblay et al. showed benefits for object detection using domain randomization\cite{tremblay2018training}. More recently, Vanherle et al. developed a Gaussian Splatting based pipeline specifically for generating high-quality, context-aware, and varied synthetic data for instance segmentation, emphasizing its role in creating diverse training instances and reducing the domain gap\cite{vanherle2025cut}. Furthermore, studies by Eli et al. on face parsing\cite{friedman2023knowing} and Jordan et al. in zero-shot classification highlight that the diversity of content within synthetic datasets can be even more crucial for model generalization than achieving perfect photorealism, a notion that resonates with our findings where varied synthetic chicken carcass data proved beneficial\cite{shipard2023boosting}.

The superior performance of the YOLOv11-seg model in our experiments, consistently achieving the highest accuracy metrics, can be attributed to its advanced architecture\cite{khanam2024yolov11}, which effectively combines efficient object detection with segmentation capabilities. Its pre-training step might also contribute to its robust feature extraction. This observation aligns with the general trend in the YOLO family, where newer iterations consistently push performance boundaries in detection and segmentation tasks, as documented in various reviews and benchmark studies on YOLO architecture evolution\cite{sapkota2024comparing}.

Furthermore, the ability of the Mask R-CNN model with an R101 backbone effectively leverage larger proportions of synthetic data compared to its R50 counterpart suggests that backbone capacity plays a significant role. A deeper network like R101 has a higher model capacity, allowing it to learn more complex features and patterns from a larger and more diverse dataset, such as that augmented with synthetic images, without saturating as quickly. This resonates with fundamental concepts in deep learning \cite{he2016deep}. More recent research by Yu et al. in the context of domain adaptation for remote sensing image classification also found that ResNet-101 achieved higher average accuracy gains (5.22\%) compared to ResNet-50 (4.99\%) when adapting to new domains, suggesting deeper architectures can indeed better utilize information from varied data distributions\cite{liang2025low}.

The substantial relative performance gains exhibited by Mask2Former when trained with augmented datasets also warrant specific discussion. This finding is likely linked to its transformer-based architecture. Transformer models are often characterized as being ``data-hungry", and their attention mechanisms may particularly benefit from the increased variability and quantity of training instances offered by synthetic data. This allows them to learn more robust and generalizable representations. This observation is echoed by broader research into transformer architectures in computer vision, such as the Vision Transformer (ViT) proposed by Dosovitskiy et al., which demonstrated that transformers can achieve remarkable performance, especially when pre-trained on large-scale datasets and subsequently fine-tuned on diverse data\cite{dosovitskiy2020image}.

For the several models tested, an optimal balance of synthetic to real data was identified, beyond which the segmentation performance plateaued or, in some isolated cases, slightly decreased. This observation suggests a critical trade-off. While synthetic data provides valuable diversity, an overreliance on it can be counterproductive, especially when a noticeable ``domain gap" exists between the synthetic and real images. In such cases, two distinct problems can arise. First, the model may begin to overfit the synthetic data, learning to recognize rendering artifacts or other patterns that are unique to the artificial images but are not present in real-world scenarios. Second, the sheer volume of synthetic examples can dilute the influence of the limited real data during training, causing the model to overlook or ``forget" subtle but crucial features that are only present in the authentic images. This nuanced behavior, where our specific models like YOLOv11-seg and Mask2Former showed sustained benefits up to a certain tested point, contrasts with some studies that might report more rapidly diminishing returns from synthetic data. Recent research supports this observation. For instance, Regina et al. in their study on drone detection, found that fine-tuning models pre-trained on synthetic data with small shares (5-10\%) of real data significantly boosts performance, highlighting the importance of real data as an ``anchor"\cite{dieter2023quantifying}. Chang et al. found that for multi-object tracking, 60-80\% of real data could be substituted with synthetic data without performance loss, emphasizing the role of flexible data generators in narrowing domain gaps\cite{chang2024equivalency}. Conversely, relying too heavily on synthetic data, especially if it is not perfectly aligned with the target domain or lacks sufficient diversity, can be detrimental for the model performance. Li et al. explicitly stated that directly using synthetic data from diffusion models can degrade performance due to feature distribution discrepancies\cite{li2025one}, and Tremblay et al. mentioned that low-quality synthetic samples can impede learning\cite{tremblay2018training}. Research into language models by Ilia et al. also warns of ``model collapse" when recursively training on purely synthetic data, suggesting a maximal proportion of synthetic data is advisable when mixed with real data to avoid degradation\cite{shumailov2024ai}. This underscores that the optimal data blend and the tolerance for synthetic data can be highly model-specific and task-dependent, influenced by factors such as model architecture, capacity, the quality and diversity of the synthetic data, and the nature of the synthetic-to-real domain gap. 

These model-specific insights into leveraging synthetic data are particularly pertinent given the well-documented challenges in data acquisition and annotation within the field of poultry. Our findings offer a potential mitigation strategy for these challenges, aligning with the successful application of synthetic data in related agricultural domains, such as precision agriculture and food processing. By demonstrating the effective use of synthetic data for chicken instance segmentation, our work contributes to the growing body of evidence supporting its role in advancing AI applications in agriculture where data scarcity is a common bottleneck.

\section{LIMITATION AND FUTURE WORK}

Despite the promising results, this study has several limitations that warrant acknowledgment.
First, the size of our real dataset (60 images for training) is quite small. While this highlights the utility of synthetic data in data-scarce scenarios, it might also exaggerate the perceived impact of synthetic augmentation, as the baseline performance with only real data is likely to be sub-optimal.
Second, the realism and specific characteristics of our synthetic data, while designed to be diverse, are still an approximation of real-world variability. There might be inherent biases in the generation process, or it may lack certain subtle differences present in real images, potentially creating a domain gap that could limit performance on truly unseen real-world data.
Third, our investigation was limited to a specific set of instance segmentation models. Other architectures, including different YOLO variants or emerging segmentation models, might respond differently to synthetic data augmentation.
Finally, our evaluation was conducted on a specific test set derived from a similar environment as the real training data. The generalization capability of the trained models to completely different plant environments remains to be thoroughly tested.

Several avenues for future research from our findings and limitations: 1)
Advanced Synthetic Data Generation: investigate more sophisticated synthetic data generation techniques, such as Generative Adversarial Networks (GANs), diffusion models or physics-based simulation, which could produce even more realistic and diverse training samples, potentially reducing the domain gap.
2) Real-World Robustness Testing: Conduct extensive testing of the trained models in diverse, real-world chicken processing line environments to assess their generalization capabilities and practical deployment readiness. 3) Exploring Synthetic-to-Real Ratios: Perform a more granular analysis of the impact of different synthetic-to-real data ratios, potentially developing adaptive strategies for determining the optimal mix for specific model architectures and dataset characteristics. 4) Cost-Benefit Analysis: Undertake a formal cost-benefit analysis comparing the resources required for generating and curating high-quality synthetic data versus the cost and effort of acquiring and annotating more real data. 5)
Domain Adaptation Techniques: Explore unsupervised or semi-supervised domain adaptation techniques to further bridge the gap between the synthetic and real data domains, aiming to improve model performance when real labeled data is extremely scarce.

\section*{Acknowledgements}
This work was supported by awards no. 2023-67021-39072, 2023-67022-39074, and 2023-67022-39075 from the U.S. Department of Agriculture (USDA)’s National Institute of Food and Agriculture (NIFA) in collaboration with the National Science Foundation (NSF) through the National Robotics Initiative (NRI) 3.0. This research is also supported by the Arkansas High Performance Computing Center, which is funded through multiple National Science Foundation grants and the Arkansas Economic Development Commission.

\bibliographystyle{IEEEtran}
\bibliography{references}

\vfill

\end{document}